\documentclass[10pt,twocolumn,letterpaper]{article}

\usepackage{iccv}
\usepackage{times}
\usepackage{epsfig}
\usepackage{graphicx}
\usepackage{amsmath}
\usepackage{amssymb}
\usepackage{breqn}
\usepackage{pgfplots}
\pgfplotsset{compat=1.18}
\usepgfplotslibrary[groupplots]

\usepackage[utf8]{inputenc} % allow utf-8 input
\usepackage[T1]{fontenc}    % use 8-bit T1 fonts
\usepackage{url}            % simple URL typesetting
\usepackage{booktabs}       % professional-quality tables
\usepackage{amsfonts}       % blackboard math symbols
\usepackage{nicefrac}       % compact symbols for 1/2, etc.
\usepackage{microtype}      % microtypography
\usepackage{tikz}        % colors
\usepackage{tabularx}
\usepackage{makecell}
\usepackage{sidecap}

\usepackage{bm}
\usepackage{tabu}
\usepackage{multirow}
\usepackage{graphics}
\usepackage{caption}
\usepackage{wrapfig}
\usepackage{enumitem}
\usepackage{subcaption}
\usepackage{arydshln}
\usepackage{algpseudocode}
\usepackage{color, colortbl}
\definecolor{Gray}{gray}{0.95}
\definecolor{orange}{rgb}{0.9,0.5,0}
\def\GeLU{\textsf{GeLU}} %normal distribution symbol
\def\DWConv{\textsf{DWConv}} %normal distribution symbol
\def\PWConv{\textsf{PWConv}} %normal distribution symbol
\def\AvgPool{\textsf{Avg-Pool}} %normal distribution symbol

\newcommand{\beq}{\vspace{0mm}\begin{equation}}
\newcommand{\eeq}{\vspace{0mm}\end{equation}}
\newcommand{\beqs}{\vspace{0mm}\begin{eqnarray}}
\newcommand{\eeqs}{\vspace{0mm}\end{eqnarray}}
\newcommand{\barr}{\begin{array}}
\newcommand{\earr}{\end{array}}

\newcommand{\Gmat}{{\bf G}}

\newcommand{\Mmat}{{\bf M}}

\newcommand{\Xmat}[0]{{{\bf X}}}

\newcommand{\Zmat}{{\bf Z}}

\newcommand{\gv}[0]{{\boldsymbol{g}}}

\newcommand{\xv}{\boldsymbol{x}}
\newcommand{\yv}{\boldsymbol{y}}
\newcommand{\zv}{\boldsymbol{z}}

\newcommand{\R}{\mathbb{R}}

\usepackage{color, colortbl}
\definecolor{Gray}{gray}{0.93}

\usepackage{xcolor,amsmath}
\usepackage[linesnumbered,ruled,vlined]{algorithm2e}
\DontPrintSemicolon

\SetKwComment{Comment}{\color{green!50!black}\# }{}

\SetKwProg{Function}{def}{:}{}

\SetKwProg{For}{for}{:}{}
\SetKwProg{If}{if}{:}{}
            
\usepackage{pifont}% http://ctan.org/pkg/pifont

\usepackage[square, sort, numbers]{natbib}

\usepackage[pagebackref=true,breaklinks=true,letterpaper=true,colorlinks,bookmarks=false]{hyperref}

\newcommand\extrafootertext[1]{%
    \bgroup
    \renewcommand\thefootnote{\fnsymbol{footnote}}%
    \renewcommand\thempfootnote{\fnsymbol{mpfootnote}}%
    \footnotetext[0]{#1}%
    \egroup
}

\iccvfinalcopy % *** Uncomment this line for the final submission

\newcommand{\txt}[1]{{\texttt{#1}}}
\def\iccvPaperID{12436} % *** Enter the ICCV Paper ID here

\ificcvfinal\fi

\usepackage[misc]{ifsym}
\newcommand{\corrAuthor}{$^{\small \textrm{\Letter}}$}

\begin{document}

%%%%%%%%% TITLE
\title{Video-FocalNets: Spatio-Temporal Focal Modulation for Video Action Recognition}

\author{%
  Syed Talal Wasim$^{1*}$\corrAuthor \quad 
  Muhammad Uzair Khattak$^{1*}$ \quad 
  Muzammal Naseer$^{1}$ \\ 
  Salman Khan$^{1,2}$ \quad
  Mubarak Shah$^{4}$ \quad
  Fahad Shahbaz Khan$^{1,3}$
  \vspace{0.5em} \\
  $^{1}$Mohamed bin Zayed University of AI \quad 
  $^{2}$Australian National University \\ 
  $^{3}$Link\"{o}ping University \quad 
  $^{4}$University of Central Florida \quad
}

\maketitle
% Remove page # from the first page of camera-ready.
\ificcvfinal\thispagestyle{empty}\fi

%%%%%%%%% ABSTRACT
\begin{abstract}
Recent video recognition models utilize Transformer models for long-range spatio-temporal context modeling. Video transformer designs are based on self-attention that can model global context at a high computational cost. In comparison, convolutional designs for videos offer an efficient alternative but lack long-range dependency modeling. Towards achieving the best of both designs, this work proposes Video-FocalNet, an effective and efficient architecture for video recognition that models both local and global contexts. Video-FocalNet is based on a spatio-temporal focal modulation architecture that reverses the interaction and aggregation steps of self-attention for better efficiency. Further, the aggregation step and the interaction step are both implemented using efficient convolution and element-wise multiplication operations that are computationally less expensive than their self-attention counterparts on video representations. We extensively explore the design space of focal modulation-based spatio-temporal context modeling and demonstrate our parallel spatial and temporal encoding design to be the optimal choice. Video-FocalNets perform favorably well against the state-of-the-art transformer-based models for video recognition on five large-scale datasets (Kinetics-400, Kinetics-600, SS-v2, Diving-48, and ActivityNet-1.3) at a lower computational cost.  Our code/models are released at \href{https://github.com/TalalWasim/Video-FocalNets}{https://github.com/TalalWasim/Video-FocalNets}. 
\end{abstract}

\extrafootertext{\textsuperscript{*}Joint first authors.\\
\indent\indent \textsuperscript{\small \corrAuthor}\txt{wasimtalal@gmail.com}}

%%%%%%%%% BODY TEXT
\section{Introduction}
% \begin{SCfigure}
% \centering
%     \centering\resizebox{\columnwidth}{!}{
%     \input{iccv2023AuthorKit/includes/figures_main/intro_graph.tex}}
%     \caption{Ablations for the number of text prompts ($M_c = 8/16/32$) and type of text prompts: Unified Context (UC) vs. Class-Specific Context (CSC) on K400. Vision prompting is fixed to global video-level prompting $G$ ($M_v = 8$) with local frame-level prompting $L$ and summary token $S$.}
%     \label{fig:intro:tradeoff}
% \end{SCfigure}

% \begin{figure}
% \centering
%     \centering
%     \input{iccv2023AuthorKit/includes/figures_main/intro_graph.tex}
%     \caption{Write caption}
%     \label{fig:intro:tradeoff}
% \end{figure}

\begin{figure}
\centering
    \includegraphics[trim={0.3cm 0.1cm 0 0},clip,width=1\columnwidth]{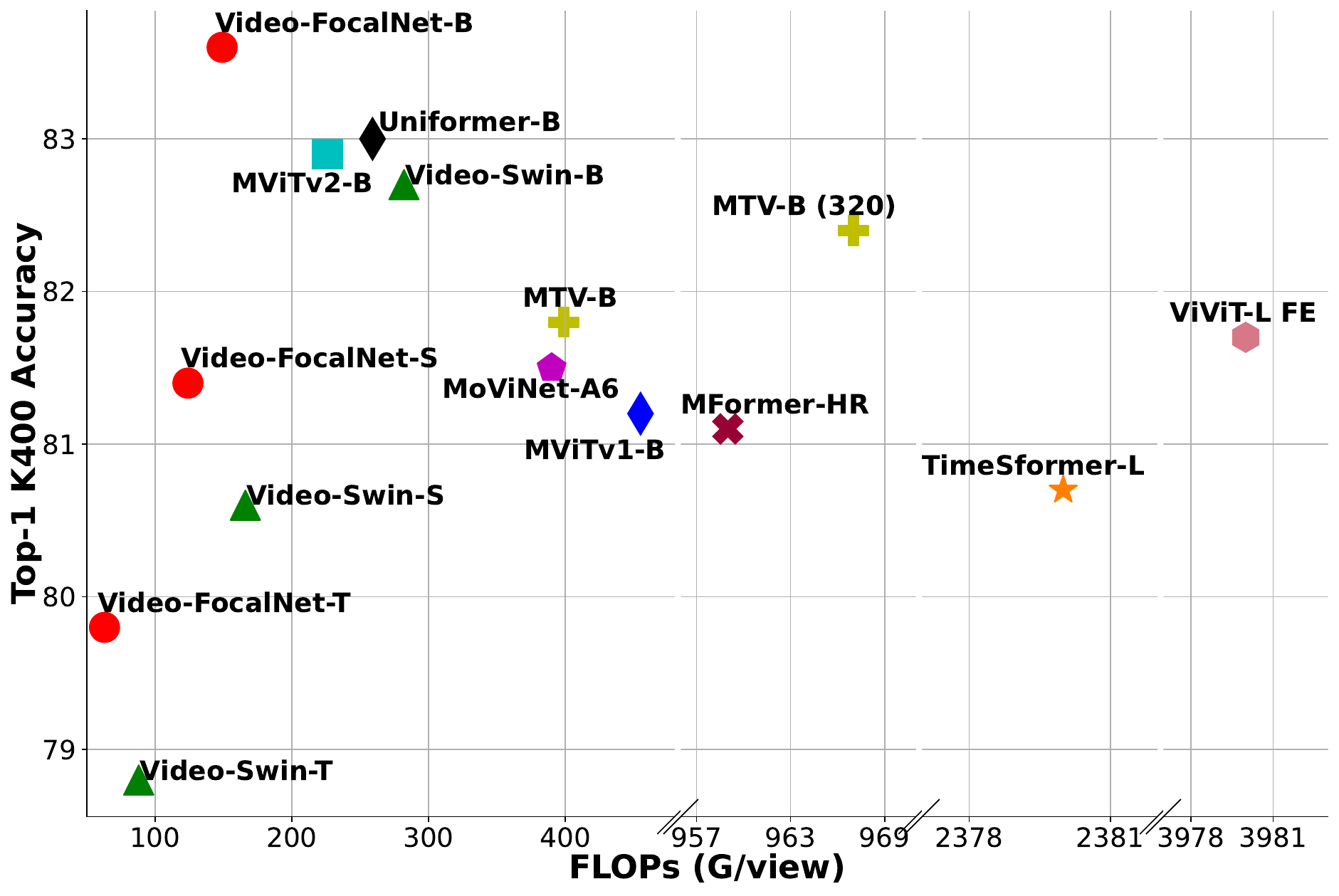}
    \caption{\emph{Accuracy vs Computational Complexity trade-off comparison:}  We show the performance of Video-FocalNets against recent methods for video action recognition. Accuracy is compared on the Kinetics-400~\cite{kay2017k400} dataset against GFLOPs/view. Our Video-FocalNets perform favorably compared to their counterparts across a range of model sizes (Tiny, Small, and Base). %Video-FocalNet-B performs better than all comparable models while Video-Focal-T and Video-FocalNet-T remain competitive against all and better than comparable methods.
    }
    \label{fig:intro:tradeoff}
\end{figure}
State-of-the-art video recognition methods have been significantly influenced by Convolutional Neural Networks (CNNs) since the introduction of Alexnet~\cite{alex2012alexnet}.
Initially 2D~\cite{karpathy2014large, Joe2014beyond, simonyan2014two} and later 3D~\cite{carreira2017quo, feichtenhofer2016spatiotemporal, tran2015learning} CNNs achieved better performance on both small-scale~\cite{kuehne2011hmdb, soomro2012ucf} and large-scale~\cite{kay2017k400, carreira2019k700, goyal2017ssv2} video recognition benchmarks. 
With their local connectivity and translational equivariance properties, CNNs have a better inductive bias especially useful for learning on small datasets. 
However, CNNs are limited in their ability to model long-range dependencies due to their limited receptive field.
On the other hand, Vision Transformers (ViTs)~\cite{dosovitskiy2021vit} offer long-range context modeling and have been quite effective for image classification~\cite{dosovitskiy2021vit, liu2021Swin, liu2022swinv2} and video recognition~\cite{gberta2021timesformer, arnab2021vivit, liu2021video-swin, shen2022mtv}. 
ViTs are based on the self-attention~\cite{vaswani2017attention} mechanism originally proposed in Natural Language Processing (NLP) that encodes minimal inductive biases and can model both short- and long-range dependencies. 
This allows ViTs to better generalize to large datasets, as shown by recent results on major video recognition benchmarks~\cite{kay2017k400, carreira2019k700, goyal2017ssv2} where they have out-performed their CNN counterparts. However, ViTs come at a high computational and parameter cost~\cite{yang2022focal}.

Video recognition requires both short-range and long-range spatio-temporal dependencies to be accurately modeled in order to achieve high performance. However, existing methods demonstrate a trade-off between efficiency and performance. While CNNs are more efficient and suited for short-range information modeling, they are limited in their representation learning capabilities for long-range dependencies and larger datasets. ViTs resolve these issues but at an increased parametric complexity and high computational cost. The high complexity originates from the dual-step self-attention operation that first performs a query-key \emph{interaction}, followed by an \emph{aggregation} over the context values. The query-key interaction requires the computationally expensive step of calculating token-to-token attention scores via dot-product since the queries and keys do not contain information about the surrounding context (they are simply linear projections of the input tokens).
In this context, this work seeks to optimize efficiency and performance while modeling both local and global contexts in videos. 

We present an effective and efficient architecture for video recognition named Video-FocalNet (\autoref{fig:intro:tradeoff}). Video-FocalNet proposes a spatio-temporal focal modulation architecture 
%based on the depthwise and pointwise convolution operators 
that reverses the steps of the self-attention operation for better efficiency. This architecture is inspired by focal modulation~\cite{yang2022focal} for image recognition and extends it to videos by independently \emph{aggregating} the surrounding spatial and temporal context for each token into spatial and temporal modulators, followed by fusing them with the queries in the \emph{interaction} step. The aggregation is based on a hierarchical contextualization step using a stack of depthwise and pointwise convolutions for the spatial and temporal branches, respectively, followed by a gated aggregation that enables modeling both short- and long-range dependencies. The aggregation step (based on depthwise/pointwise convolutions) and the interaction step (based on element-wise multiplication) are both computationally less expensive than their self-attention counterparts i.e., query-key interactions and query-value aggregation via matrix multiplications. 

We extensively explore various design configurations for optimal spatio-temporal context modeling with focal modulation. Our analysis shows that the proposed parallel spatial and temporal focal modulation design offers the best performance and is suitably efficient compared to other sequential designs. We introduce a family of Video-FocalNet architectures (tiny, small, and base) based on spatio-temporal focal modulation and demonstrate their favorable performance compared to state-of-the-art transformer-based methods on video recognition at a lower computational cost.
Our major contributions are summarized as follows:\vspace{-0.3em}
\begin{itemize}\setlength{\itemsep}{0em}
    \item We tackle the challenge of effective spatio-temporal modeling for video recognition. To solve this challenge, we propose a video-focal modulation block that is able to use computationally efficient depthwise and pointwise convolutions through a hierarchical context aggregation design for local-global context modeling.
    \item We explore various design choices for spatio-temporal focal modulation and propose a parallel design for spatial and temporal encoding that optimizes for both performance and computation cost as shown in \autoref{fig:ablation:design_tradeoff}.
    \item We achieve state-of-the-art performance on three major benchmarks: Kinetics-400~\cite{kay2017k400}, Kinetics-600~\cite{carreira2018k600} and Something-Something-v2~\cite{goyal2017ssv2}, surpassing comparable methods in literature by $0.6\%$, $1.2\%$ and $0.6\%$ respectively. Also, we outperform previous works on the relatively smaller Diving-48~\cite{yingwei2018d48} and ActivityNet-1.3~\cite{heilbron2015anet} datasets. We achieve an optimal trade-off between accuracy and computation cost as shown in \autoref{fig:intro:tradeoff}.
\end{itemize}

%-------------------------------------------------------------------------

\section{Related Work}

\textbf{Video Recognition:}
Early methods in video recognition are feature-based ~\cite {klaser2008spatio, laptev2005space, wang2013dense}. However, with the startling success of 2D CNNs~\cite{alex2012alexnet, simonyan2014vgg, he2016resnet, tan2019efficientnet} on Imagenet~\cite{deng2009imagenet}, they were also introduced to the task of video recognition~\cite{karpathy2014large, Joe2014beyond, simonyan2014two}. Later, after the release of large-scale datasets, such as Kinetics~\cite{kay2017k400}, the 3D CNN-based methods were introduced~\cite{carreira2017quo, feichtenhofer2016spatiotemporal, tran2015learning}. These were much more effective in modeling spatio-temporal relations and outperformed 2D CNN-based methods. However, the computational cost for these 3D CNN-based methods was quite prohibitive. Therefore, various variants of the 3D CNNs were introduced~\cite{feichtenhofer2020expanding, sun2015human, szegedy2016rethinking, tran2018closer, xie2018s3d, li2020tea, lin2019tsm, qiu2019learning, feichtenhofer2019slowfast, duan2020omni}, which decreased computation cost and improved performance. With the success of the Vision Transformer~\cite{dosovitskiy2021vit} for image recognition, they were also introduced to video recognition. The first methods in this area used a combination of Vision Transformers and CNNs~\cite{wang2018nonlocal, wang2020nas, kondratyuk2021movinets}, including transformer blocks to model the longer range context. Later advancements then introduced fully transformer based architectures~\cite{liu2021video-swin, arnab2021vivit, gberta2021timesformer, shen2022mtv, zhang2021vidtr, patrick2021keeping, fan2021multiscale, li2022improved}, which outperformed all previous methods across multiple benchmarks. Recently, a new method~\cite{li2022uniformer} has been proposed, combining CNNs and ViTs, which achieves comparable performance to state-of-the-art fully transformer-based methods.

\textbf{Global Context Modeling:}
Due to the localized nature of 2D CNNs, global context modeling was lacking in pure 2D CNN-based computer vision methods. Self-attention~\cite{vaswani2017attention} was introduced to model long-range dependencies for visual inputs. However, self-attention comes at a high computation cost due to the required matrix multiplications. Various approaches have been introduced to address this problem. These include local window based attention~\cite{liu2021Swin, liu2022swinv2, zhang2021longformer, chu2021twins, yang2021focalattn, parmar2018image, qiu2019blockwise}, along with variants that add global tokens to model global information~\cite{ainslie2020etc, beltagy2020long, jaegle2021perceiver, zaheer2021bigbird, ni2022expanding}. To reduce the computation cost, some methods used computationally efficient patterns for attention such as strided~\cite{child2019generating} and axial~\cite{ho2019axial} patterns, as well as attention computed along the channel dimension rather than the token dimension~\cite{el-nouby2021xcit}. Other methods also combined convolution and self-attention for local and global modeling~\cite{xu2021co-scale, wu2021cvt, guo2021cmt, li2021localvit}. Various methods for linearizing self-attention were also investigated, including the projection of token dimensions~\cite{wang2020linformer, kitaev2020reformer}, factorizing the softmax-attention kernel~\cite{choromanski2020rethinking, shen2021efficient, xiong2021nyströmformer}. Using CNNs, a new method for modeling global context, termed focal modulation \cite{yang2022focal}, has been proposed recently. To model local and global information, focal modulation employs hierarchical context aggregation to combine information from increasing sizes of receptive fields.

%-------------------------------------------------------------------------

\section{Methodology}
\begin{figure*}
\centering
    \includegraphics[width=\textwidth]{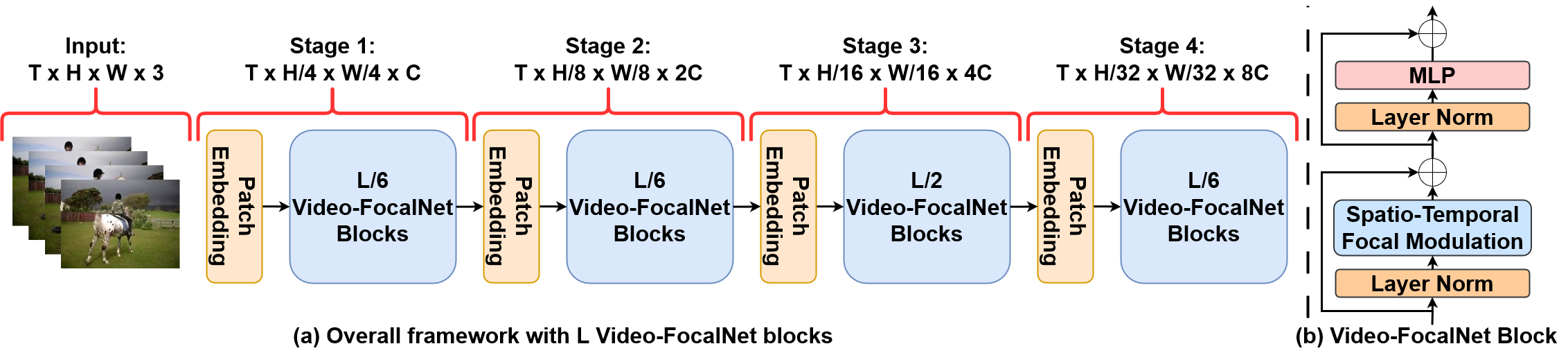}
    \caption{\textbf{(a)} \textbf{The overall architecture of Video-FocalNets:} Following~\cite{liu2021Swin, yang2021focalattn, yang2022focal}, we define a four-stage architecture, with each stage comprising a patch embedding and a number of Video-FocalNet blocks. The total number of blocks is $L$, with stages one, two, three, and four having $L/6$, $L/6$, $L/2$, and $L/6$ blocks respectively. \textbf{(b)} \textbf{Single Video-FocalNet block:} Similar to the transformer blocks~\cite{vaswani2017attention}, we replace self-attention with Spatio-Temporal Focal Modulation.}
    \label{fig:main_arch:overall_architecture}
\end{figure*}

Let us assume a video input is encoded to produce a feature representation $\Xmat_{st} \in \R^{T \times H \times W \times C}$ with $T$ frames, $H\times W$ spatial resolution, and $C$ channels respectively. To obtain the spatio-temporal context enriched representation $\yv_i \in \R^{C}$ for a given token (query) $\xv_i \in \R^{C}$ in the input spatio-temporal feature map $\Xmat_{st}$, it is necessary to perform an \emph{interaction} between the query and its neighboring spatial and temporal tokens, and then \emph{aggregate} the resulting information over the surrounding spatio-temporal contexts. To effectively model a spatio-temporal input, it is important to encode both short-range and long-range dependencies for the enriched context modeling for videos.

Self-attention~\cite{vaswani2017attention} which is used in state-of-the-art video recognition methods~\cite{liu2021video-swin, arnab2021vivit, gberta2021timesformer, shen2022mtv, zhang2021vidtr, patrick2021keeping, fan2021multiscale}, uses a First Interaction, Last Aggregation (FILA) process which involves initially calculating the attention scores through the query and key interaction $\mathcal{T}_1$, followed by aggregation $\mathcal{M}_1$ over the contexts as shown in \autoref{Eq:late_agg}.
\begin{equation}
% \small
    \yv_i = \mathcal{M}_1 ( \mathcal{T}_1 (\xv_i, \Xmat_{st}),  \Xmat_{st} ).  \hspace{-0mm}
    \label{Eq:late_agg}
\end{equation}
Since the query and keys during the interaction process are simple linear projections of the input feature map, self-attention involves computationally expensive token-to-token attention score calculation through query-key interactions because individual keys do not contain information about the surrounding context.

Recently, a new encoding method, Focal modulation~\cite{yang2022focal} has been proposed, which follows an early aggregation process by First Aggregation, Last Interaction (FALI) mechanism. Essentially, both self-attention and focal modulation involve the \emph{interaction} and \emph{aggregation} operations but differ in the sequence of operation. In the case of focal modulation, context aggregation $\mathcal{M}_2$ is performed first, followed by the interaction $\mathcal{T}_2$ between the queries and the aggregated features, as shown in \autoref{Eq:visual_modulation}.
\begin{equation}
% \small
    \yv_i = \mathcal{T}_2 ( \mathcal{M}_2 (i, \Xmat_{st}),  \xv_i ). \hspace{0mm}
    \label{Eq:visual_modulation}
\end{equation}
The output of the aggregation is known as the \emph{modulator} which encodes the surrounding context for each query. Note that the operator $\mathcal{M}_2$ in focal modulation is based on convolutions, which are computationally more efficient compared to $\mathcal{M}_1$ in self-attention. Similarly, the interaction operator $\mathcal{T}_2$ is a simple element-wise multiplication, compared to the token-to-token attention score computation in self-attention which has quadratic complexity.

The focal modulation process given by~\cite{yang2022focal} works well on images by extracting the \emph{spatial} context around a query token. However, to model spatio-temporal information, both the \emph{spatial} and \emph{temporal} contexts surrounding a single query token have to be extracted. To achieve this we propose our architecture, Video-FocalNets which explicitly models both intra-frame (spatial) and inter-frame (temporal) information. Our approach aims to independently model the spatial and temporal information by proposing a two-stream spatio-temporal focal modulation block, in which one branch learns spatial information and the other models the temporal information. By decoupling the spatial and temporal branches, we are able to separately extract and aggregate spatial and temporal context for each query token, generating spatial and temporal \emph{modulators}. These modulators are then fused with the query tokens to build the final feature map.

Our design transfers the desirable qualities of late aggregation in focal modulation to the video tasks. Particularly, Focal Modulation is performed for each target token with the context centered around it, hence it is translationally invariant. It also decouples the queries from the context around them, allowing the queries to preserve fine-grained information, while the coarser context surrounding it is extracted. Focal modulation uses a hierarchical-gated aggregation method, to aggregate information across multiple levels of granularity. This allows for modeling both short- and long-range dependencies within a video while improving computational and parameter efficiency.

We now present our approach on spatio-temporal focal modulation in \autoref{sec:meth:vfn}, specifying the \emph{Hierarchical Contextualization} and \emph{Gated Aggregation} processes for videos in \autoref{sec:meth:focal:STHC} and \autoref{sec:meth:focal:STGA}, respectively. For consistency, we maintain the same terminology as proposed in~\cite{yang2022focal}. Finally, we outline our network architecture variants in \autoref{sec:meth:nav}.

\begin{figure}[t]
    \centering
    \includegraphics[width=1\columnwidth]{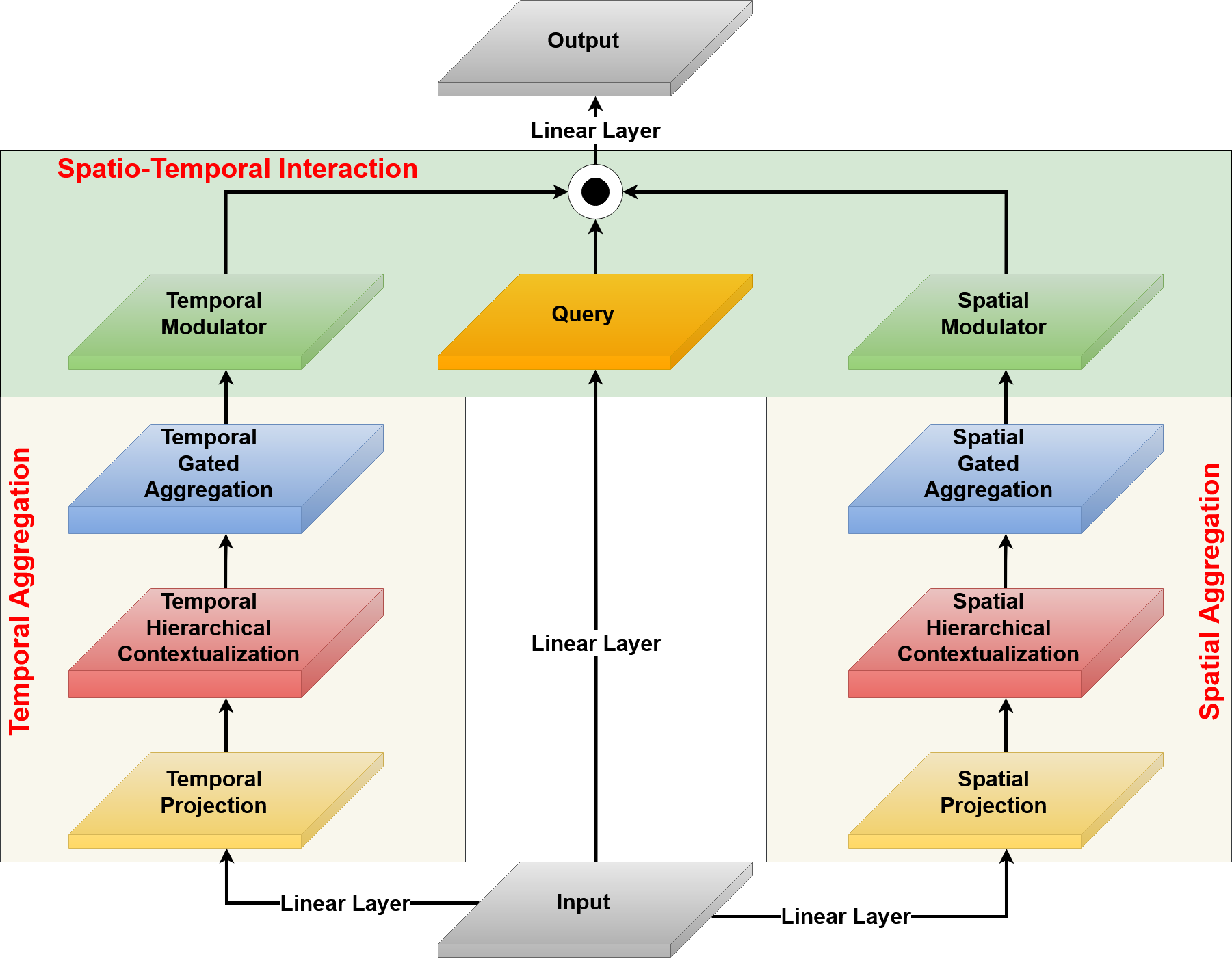}
    \caption{\textbf{The Spatio-Temporal Focal Modulation layer:} We design a spatio-temporal focal modulation block that independently models the spatial and temporal information.
    The input is first projected using linear layers to produce queries, spatial/temporal feature maps, and spatial/temporal gates. Then through hierarchical contextualization ($\mathcal{M}_s/\mathcal{M}_t$) and gated aggregation ($\Gmat_s/\Gmat_t$), the spatial and temporal modulators are produced. These then interact with the query tokens through element-wise multiplication operation (\protect\includegraphics[height=0.3cm]{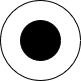}) to form the final spatio-temporal feature map.}
    \label{fig:main_arch:overview_focal_modulation}
\end{figure}

\subsection{Spatio-Temporal Focal Modulation}
\label{sec:meth:vfn}

To model the spatial and temporal dimensions, we propose a two-stream spatio-temporal focal modulation block. The overall architecture is presented in \autoref{fig:main_arch:overall_architecture} and the design of the spatio-temporal focal modulation is presented in \autoref{fig:main_arch:overview_focal_modulation}. We validate its effectiveness via detailed ablations and comparisons with alternate design choices in \autoref{sec:exp:ablations}. For an input spatio-temporal feature map $\Xmat_{st} \in \R^{T \times H \times W \times C}$, the two-stream spatio-temporal encoding process involves independent aggregations along the spatial and temporal dimensions, followed by a joint interaction with the queries, as shown in \autoref{Eq:spt_visual_modulation}.
\begin{equation}
% \small
    \yv_i = \mathcal{T}_{st} ( \mathcal{M}_s (i_{t}, \Xmat_{st,t}),  \mathcal{M}_t (i_{hw}, \Xmat_{st,hw}), \xv_i), \hspace{0mm}
    \label{Eq:spt_visual_modulation}
\end{equation}
where $\Xmat_{st,t} \in \R^{H \times W \times C}$ is a single spatial slice for the temporal dimension $t \in \{1,...,T\}$, while $i_{t}$ is the spatial location for the slice $t \in \{1,...,T\}$. Similarly, $\Xmat_{st,hw} \in \R^{T \times C}$ is a single temporal slice for the spatial dimensions $h \in \{1,...,H\}$ and $w \in \{1,...,w\}$, while $i_{hw}$ is the temporal location. The operators $\mathcal{M}_s$ and $\mathcal{M}_t$ are based on the depth-wise convolution and point-wise convolution operators respectively and $\mathcal{T}_{st}$ is an element-wise multiplication. The spatio-temporal focal modulation process can therefore be defined as follows:
\begin{equation}
% \small
    \yv_i = q(\xv_i) \odot m_s (i_t, \Xmat_{st,t}) \odot m_t (i_{hw}, \Xmat_{st,hw}),  \hspace{-5mm}
    \label{Eq:spt_FocalModulationSummary}
\end{equation}
where $q(\cdot)$ is a query projection function and $\odot$ is the element-wise multiplication, $m_s(\cdot)$ and $m_t(\cdot)$ are context aggregation functions, whose outputs are called \emph{spatial modulator} and \emph{temporal modulator} respectively. The formulation of $m_s(\cdot)$ and $m_t(\cdot)$ involves two steps: \emph{Hierarchical Contextualization} and \emph{Gated Aggregation}. The following \autoref{sec:meth:focal:STHC} and \autoref{sec:meth:focal:STGA} talk about \emph{Spatio-Temporal Hierarchical Contextualization} and \emph{Spatio-Temporal Gated Aggregation} respectively.

\subsubsection{Spatio-Temporal Hierarchical Contextualization}
\label{sec:meth:focal:STHC}

We first project the input spatio-temporal feature map $\Xmat_{st} \in \R^{T \times H \times W \times C}$ using two linear layers, producing $\Zmat^0_s$ and $\Zmat^0_t$, as defined by \autoref{eq:sthc1}.
\begin{equation}
\begin{aligned}
% \small
    \Zmat^0_s&=f_{z,s}(\Xmat_{st}) \in \R^{T \times H \times W \times C},\\
    \Zmat^0_t&=f_{z,t}(\Xmat_{st}) \in \R^{T \times H \times W \times C},
\end{aligned}
\label{eq:sthc1}
\end{equation}
where $f_{z,s}$ and $f_{z,t}$ are the spatial and temporal linear projection layers respectively. We then apply a series of $L$ depth-wise convolutions ($\DWConv$) and point-wise convolutions ($\PWConv$) to the respective spatial and temporal projected inputs $\Zmat^0_s$ and $\Zmat^0_t$ along the spatial and temporal dimensions respectively. The outputs $\Zmat^{\ell}_s$ and $\Zmat^{\ell}_t$, at each focal level $\ell \in \{1,...,L\}$, are therefore given as:
\begin{equation}
\small
\begin{aligned}
    \Zmat^{\ell}_s &= f_{a,s}^{\ell}(\Zmat^{\ell-1}_s) \triangleq \GeLU( \DWConv( \Zmat^{\ell-1}_s )) \in \R^{T \times H \times W \times C},\\
    \Zmat^{\ell}_t &= f_{a,t}^{\ell}(\Zmat^{\ell-1}_t) \triangleq \GeLU( \PWConv( \Zmat^{\ell-1}_t )) \in \R^{T \times H \times W \times C},
\end{aligned}
\label{eq:spt_hier_context}
\end{equation}
where $f_{a,s}^{\ell}(\cdot)$ and $f_{a,t}^{\ell}(\cdot)$ are the spatial and temporal contextualization functions with $\GeLU$~\cite{hendrycks2016gelu} activation function. To obtain the global representation, a global average pooling operation is performed along the spatial and temporal dimensions on $\Zmat^L_s$ and $\Zmat^L_t$ respectively as shown in \autoref{eq:spt_global_context}.
\begin{equation}
% \small
\begin{aligned}
    \Zmat^{L+1}_s = \AvgPool(\Zmat^L_s),\\
    \Zmat^{L+1}_t = \AvgPool(\Zmat^L_t),
\end{aligned}
\label{eq:spt_global_context}
\end{equation}
where $\AvgPool$ is the global average pool operator.

\begin{figure*}[t!]
\centering
    \includegraphics[width=\textwidth]{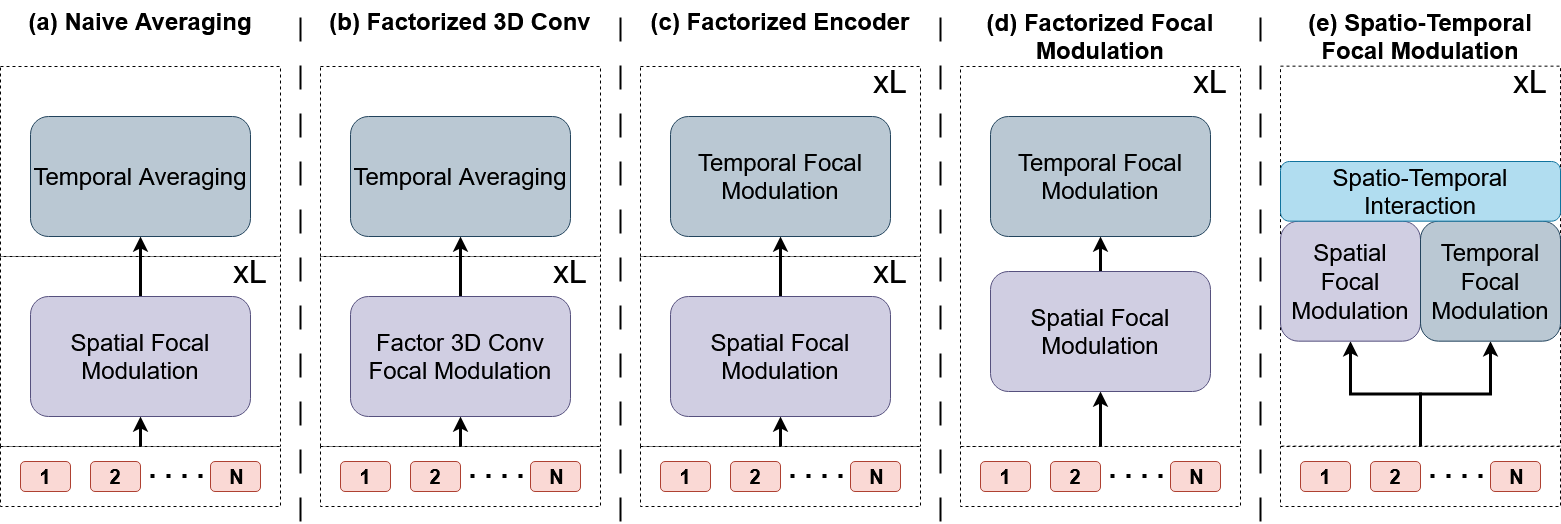}
    \caption{We show various design choices for spatio-temporal context modeling via Focal Modulation and evaluate them in \autoref{fig:ablation:design_tradeoff}. \textbf{(a)} A naive solution where the frames are passed through spatial focal modulation layers and averaged. \textbf{(b)} A variation of the naive solution replacing 2D depthwise convolution with factorized 3D convolution (2D depthwise followed by 1D pointwise convolution). \textbf{(c)} A factorized encoder design that stacks two encoders, one modeling spatial and the other the temporal dimension. \textbf{(d)} A design based on~\cite{gberta2021timesformer} which uses factorized spatial and temporal focal modulation. \textbf{(e)} Our proposed spatio-temporal focal modulation with parallel spatial and temporal branches followed by spatio-temporal interaction.}
    \label{fig:ablation:design}
\end{figure*}
\begin{figure}[t]
    \centering
    \includegraphics[width=1\columnwidth]{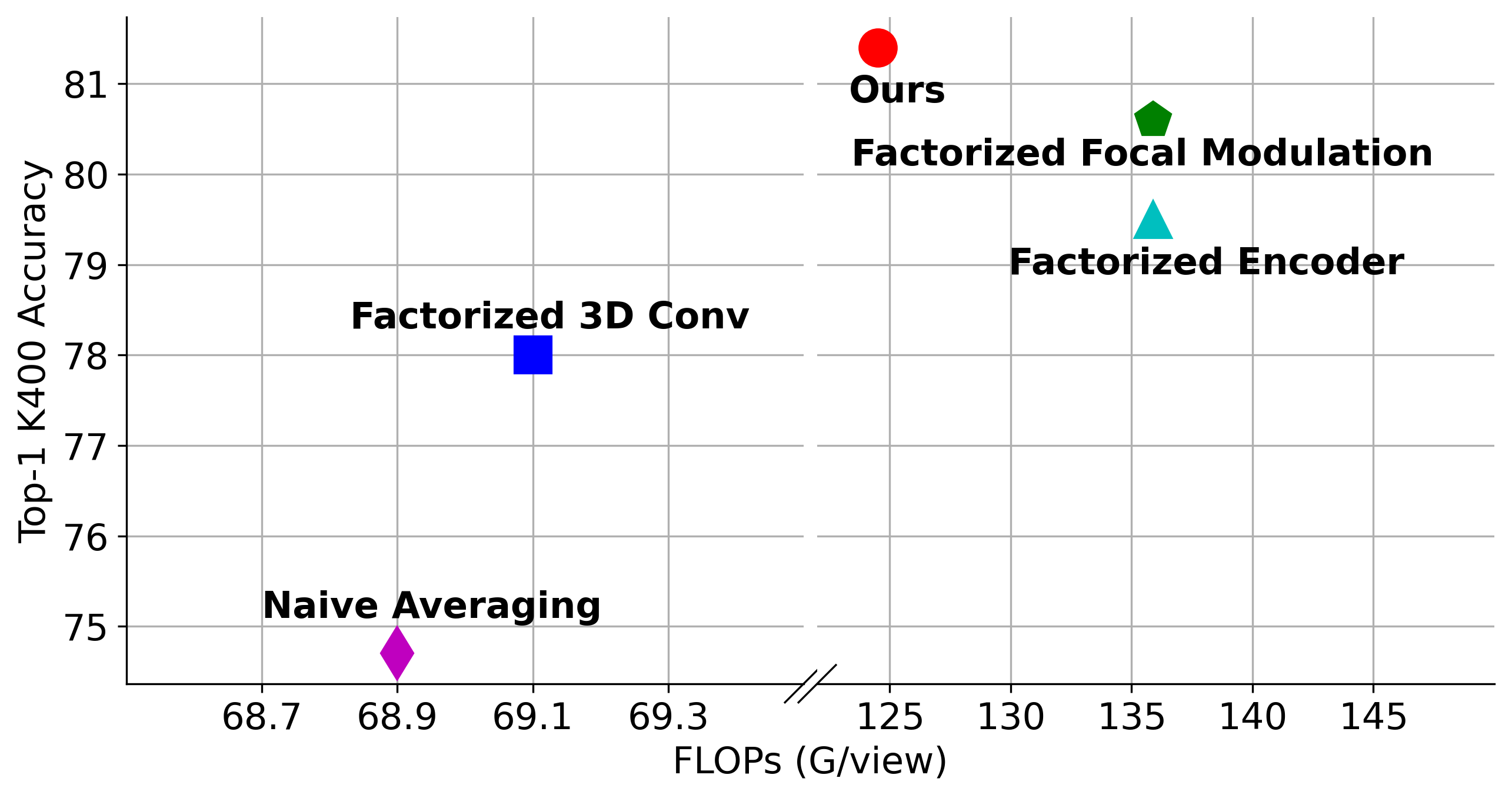}
    \caption{Comparison of various design choices for Video-FocalNet-S on Kinetics-400~\cite{kay2017k400} validation set.}
    \label{fig:ablation:design_tradeoff}
    % \vspace{-0.3cm}
\end{figure}

\subsubsection{Spatio-Temporal Gated Aggregation}
\label{sec:meth:focal:STGA}

Next, we condense the respective spatial and temporal feature maps, $\Zmat^{\ell}_s$ and $\Zmat^{\ell}_t$, into the respective spatial and temporal modulators through a gating mechanism. We obtain the respective spatial and temporal gating weights, $\Gmat_s = f_{g,s}(\Xmat_{st}) \in \R^{H \times W \times (L+1)}$ and $\Gmat_t = f_{g,t}(\Xmat_{st}) \in \R^{T \times (L+1)}$, using the linear projection layers $f_{g,s}$ and $f_{g,t}$. This is followed by a dot product between the feature maps and their respective gates, as shown in \autoref{eq:spt_gated_agg}.
\begin{equation}
\small
\begin{aligned}
    \Zmat^{out}_s & = \sum_{\ell=1}^{L+1} \Gmat_s^\ell \odot \Zmat_s^\ell \in \R^{H \times W \times C},\\
    \Zmat^{out}_t & = \sum_{\ell=1}^{L+1} \Gmat_t^\ell \odot \Zmat_t^\ell \in \R^{T \times C},
\end{aligned}
\label{eq:spt_gated_agg}
\end{equation}
where $\Zmat^{out}_s$ and $\Zmat^{out}_t$ are the single aggregated spatial and temporal feature maps and $\Gmat^{\ell}_s \in \R^{H \times W \times 1}$ and $\Gmat^{\ell}_t \in \R^{T \times 1}$ are slices of $\Gmat_s$ and $\Gmat_t$ respectively for the level $\ell$. To enable communication across different channels, another set of linear layers, $h_s(\cdot)$ and $h_t(\cdot)$, are used to obtain the \emph{spatial modulator} ($\Mmat_s = h_s(\Zmat^{out}_s) \in \R^{T \times H \times W \times C}$) and \emph{temporal modulator} ($\Mmat_t = h_t(\Zmat^{out}_t) \in \R^{T \times H \times W \times C}$) respectively.

Therefore, the spatio-temporal focal modulation process defined by \autoref{Eq:spt_FocalModulationSummary} can be rewritten as:
\begin{equation}
    \yv_i = q(\xv_i) \odot h_s(\sum_{\ell=1}^{L+1} \gv^\ell_{i,s} \cdot \zv^\ell_{i,s}) \odot h_t(\sum_{\ell=1}^{L+1} \gv^\ell_{i,t} \cdot \zv^\ell_{i,t})
    \label{Eq:spt_FocalModulation}
\end{equation}
where $\zv_{i,s}^\ell$/$\zv_{i,t}^\ell$ and $\gv_{i,s}^\ell$/$\gv_{i,t}^\ell$ are the spatial/temporal visual feature and spatial/temporal gating value at location $i$ of $\Zmat^\ell_s$/$\Zmat^\ell_t$ and $\Gmat^\ell_s$/$\Gmat^\ell_t$ respectively.

\subsubsection{Design Variations}
\label{sec:meth:design_choices}

We further compare our proposed spatio-temporal focal modulation design against various other possible designs shown in \autoref{fig:ablation:design}. This explorative study validates the proposed design to be the optimal one. The first design, \textbf{(a)}, is a simple extension of the spatial focal modulation to videos, which passes each frame through the spatial encoder (which uses only 2D depthwise convolution) and averages along the temporal dimension. Mathematically, \autoref{Eq:spt_visual_modulation} for this case can be re-written as:

\begin{equation}
% \small
    \yv_i = \mathcal{T}_{st} ( \mathcal{M}_s (i_{t}, \Xmat_{st,t})). \hspace{0mm}
    \label{Eq:spt_design_1}
\end{equation}

A variation of this design, \textbf{(b)}, uses factorized 3D convolution (2D depthwise followed by 1D pointwise convolution).

The next \textbf{(c)} uses a factorized encoder that stacks two encoders, one spatial (using 2D depthwise convolution) and one temporal (using 1D depthwise convolution), on top of each other. This is similar to the factorized encoder design presented by~\cite{arnab2021vivit} but replaces spatial and temporal self-attention with spatial and temporal focal modulation.

The second last design \textbf{(d)} follows the concept of divided space-time attention proposed by~\cite{gberta2021timesformer} and uses alternating spatial and temporal focal modulation.

The final design \textbf{(e)} is the proposed spatio-temporal focal modulation. The accuracy and computation requirements for each are reported on the Kinetics-400 dataset in \autoref{fig:ablation:design_tradeoff}. It can be seen that the proposed design is the best in terms of accuracy and computation.

\subsection{Network Variants}
\label{sec:meth:nav}

Following~\cite{liu2021video-swin, yang2022focal}, we use the same four-stage layouts and hidden dimensions as in~\cite{yang2022focal}, but replace the focal modulation block with our spatio-temporal focal modulation block. In each stage, a stack of $L$ Video-FocalNet blocks is used, divided between the four stages as $\{L/6, L/6, L/2, L/6\}$. We introduce four different versions of Video-FocalNets. The architecture hyper-parameters of these model variants are:

\begin{itemize}
    \item Video-FocalNet-T: $C=96$, block$_{\textit{num}}$ = $\{2, 2, 6, 2\}$
    \item Video-FocalNet-S: $C=96$, block$_{\textit{num}}$ = $\{2, 2, 18, 2\}$
    \item Video-FocalNet-B: $C=128$, block$_{\textit{num}}$ = $\{2, 2, 18, 2\}$
\end{itemize}

We use non-overlapping convolution layers for patch embedding at the beginning (kernel size=$4\times 4$, stride=4) and between two stages (kernel size=$2 \times 2$, stride=2), respectively. The focal levels ($L$) for the models are set to $2$ with the kernel for the first level set to $k^1 = 3$. We gradually increase the kernel size by 2 from lower focal levels to higher ones, \textit{i.e.}, $k^\ell = k^{\ell-1}+2$.

%-------------------------------------------------------------------------

\section{Results and Analysis}
\label{sec:exp}

\begin{table}[t]
\caption{Training hyperparameters for experiments in the main paper.  ``--'' indicates that the regularisation method was not used at all. Values that are constant across all columns are listed once. Datasets are denoted as follows: K400: Kinetics-400. K600: Kinetics-600. SS-v2: Something-Something-v2. D-48: Diving 48. ANet-1.3: ActivityNet-1.3.}

\setlength{\tabcolsep}{3mm}
% \scalebox{\columnwidth}[\columnwidth]{
\resizebox{\linewidth}{!}{

\begin{tabular}{lccccc} %
\toprule
                              & K400 & K600 & SS-v2 & D-48 & ANet-1.3 \\ \midrule
\multicolumn{6}{l}{\textit{Optimization}}                                                                                 \\
Optimizer                                                   & \multicolumn{5}{c}{SGD}\\
Batch size                                                  & \multicolumn{5}{c}{512} \\
Learning rate schedule	                                    & \multicolumn{5}{c}{cosine with linear warmup} \\
Linear warmup epochs	                                    & \multicolumn{5}{c}{20} \\
Base learning rate			                                & \multicolumn{5}{c}{0.1} \\
Epochs	                                                    & \multicolumn{5}{c}{120}\\
\midrule
\multicolumn{5}{l}{\textit{Data augmentation}} \\
Random crop probability                                     & \multicolumn{5}{c}{1.0} \\
Random flip probability                                     & 0.5 & 0.5 & -- & -- & -- \\
Scale jitter probability	                                & \multicolumn{5}{c}{1.0} \\
Maximum scale			                                    & \multicolumn{5}{c}{1.33} \\
Minimum scale			                                    & \multicolumn{5}{c}{0.75} \\
Colour jitter probability                                   & \multicolumn{5}{c}{0.8}  \\
\midrule
\multicolumn{5}{l}{\textit{Other regularisation}} \\
Stochastic droplayer rate~\cite{huang2016stochasticdepth} & \multicolumn{5}{c}{0.1} \\
Label smoothing~\cite{szegedy2015inception}				  & \multicolumn{5}{c}{0.1} \\
Mixup ($\alpha=0.8$) probability~\cite{zhang2018mixup}				  & \multicolumn{5}{c}{0.5} \\
\bottomrule
\end{tabular}}
\label{tab:training_hyperparameters}
\vspace{-.3cm}
\end{table}
\begin{table*}[t!]

\centering
\small
\caption{Comparison with state-of-the-art methods on Kinetics-400~\cite{kay2017k400}.}\vspace{-0.5em}
\setlength{\tabcolsep}{3mm}
\scalebox{0.9}[0.9]{
\begin{tabular}{lcccc}
\rowcolor{Gray}
    \toprule
    Method                                                               & Pre-training     & Top-1              & Views                     & FLOPs (G/view)\\
    \midrule
    TEA~\textit{(ICCV'21)}~\cite{li2020tea}                              & ImageNet-21K     & 76.1               & $10 \times 3$             & 70  \\
    TSM-ResNeXt-101~\textit{(ICCV'21)}~\cite{lin2019tsm}                 & ImageNet-21K     & 76.3               & -                         & -    \\
    I3D NL~\textit{(ICCV'21)}~\cite{wang2018nonlocal}                    & ImageNet-21K     & 77.7               & $10 \times 3$             & 359 \\
    VidTR-L~\textit{(ICCV'21)}~\cite{zhang2021vidtr}                     & ImageNet-21K     & 79.1               & $10 \times 3$             & 351 \\
    LGD-3D R101~\textit{(CVPR'19)}~\cite{qiu2019learning}                & ImageNet-21K     & 79.4               & -                         & -    \\
    SlowFast R101-NL~\textit{(ICCV'19)}~\cite{feichtenhofer2019slowfast} & ImageNet-21K     & 79.8               & $10 \times 3$             & 234  \\
    X3D-XXL~\textit{(CVPR'20)}~\cite{feichtenhofer2020expanding}         & ImageNet-21K     & 80.4               & $10 \times 3$             & 194  \\
    OmniSource~\textit{(ECCV'20)}~\cite{duan2020omni}                    & ImageNet-21K     & 80.5               & -                         & -    \\
    TimeSformer-L~\textit{(ICML'21)}~\cite{gberta2021timesformer}        & ImageNet-21K     & 80.7               & $1 \times 3$              & 2380 \\
    MFormer-HR~\textit{(NeurIPS'21)}~\cite{patrick2021keeping}           & ImageNet-21K     & 81.1               & $10 \times 3$             & 959 \\
    MViTv1-B~\textit{(ICCV'21)}~\cite{fan2021multiscale}                 & -                & 81.2               & $3 \times 3$              & 455  \\
    MoViNet-A6~\textit{(CVPR'21)}~\cite{kondratyuk2021movinets}          & ImageNet-21K     & 81.5               & $1 \times 1$              & 390  \\
    ViViT-L FE~\textit{(CVPR'21)}~\cite{arnab2021vivit}                  & ImageNet-21K     & 81.7               & $1 \times 3$              & 3980 \\
    MTV-B~\textit{(CVPR'22)}~\cite{shen2022mtv}                          & ImageNet-21K     & 81.8               & $4 \times 3$              & 399  \\
    MTV-B (320p)~\textit{(CVPR'22)}~\cite{shen2022mtv}                   & ImageNet-21K     & 82.4               & $4 \times 3$              & 967  \\
    Video-Swin-T~\textit{(CVPR'22)}~\cite{liu2021video-swin}             & ImageNet-1K      & 78.8               & $4 \times 3$              & 88  \\
    Video-Swin-S~\textit{(CVPR'22)}~\cite{liu2021video-swin}             & ImageNet-1K      & 80.6               & $4 \times 3$              & 166  \\
    Video-Swin-B~\textit{(CVPR'22)}~\cite{liu2021video-swin}             & ImageNet-1K      & 80.6               & $4 \times 3$              & 282  \\
    Video-Swin-B~\textit{(CVPR'22)}~\cite{liu2021video-swin}             & ImageNet-21K     & 82.7               & $4 \times 3$              & 282  \\
    MViTv2-B~\textit{(CVPR'22)}~\cite{li2022improved}                    & -                & 82.9               & $5 \times 1$              & 226  \\
    Uniformer-B~\textit{(ICLR'22)}~\cite{li2022uniformer}                & ImageNet-1K      & 83.0               & $4 \times 3$              & 259  \\
    \midrule
    Video-FocalNet-T                                                     & ImageNet-1K      & 79.8               & $4 \times 3$              & 63  \\
    Video-FocalNet-S                                                     & ImageNet-1K      & 81.4               & $4 \times 3$              & 124  \\
    Video-FocalNet-B                                                     & ImageNet-1K      & \textbf{83.6}      & $4 \times 3$              & 149  \\
    \bottomrule
\end{tabular}}
\label{tab:k400}
\vspace{-.3cm}

\end{table*}

\begin{table}[t!]

\centering
\small
\caption{Comparison with state-of-the-art methods on Kinetics-600~\cite{carreira2018k600} dataset.}\vspace{-0.5em}
\setlength{\tabcolsep}{3mm}
\resizebox{0.95\columnwidth}{!}{
\begin{tabular}{lcc}
\rowcolor{Gray}
    \toprule
    Method                                                               & Pre-training     & Top-1\\
    \midrule
    SlowFast R101-NL~\textit{(ICCV'19)}~\cite{feichtenhofer2019slowfast} & ImageNet-21K     & 81.8\\
    X3D-XXL~\textit{(CVPR'20)}~\cite{feichtenhofer2020expanding}         & ImageNet-21K     & 81.9\\
    TimeSformer-L~\textit{(ICML'21)}~\cite{gberta2021timesformer}        & ImageNet-21K     & 82.2\\
    MFormer-HR~\textit{(NeurIPS'21)}~\cite{patrick2021keeping}           & ImageNet-21K     & 82.7\\
    ViViT-L FE~\textit{(CVPR'21)}~\cite{arnab2021vivit}                  & ImageNet-21K     & 82.9\\
    MTV-B~\textit{(CVPR'22)}~\cite{shen2022mtv}                          & ImageNet-21K     & 83.6\\
    MTV-B (320p)~\textit{(CVPR'22)}~\cite{shen2022mtv}                   & ImageNet-21K     & 84.0\\
    Video-Swin-B~\textit{(CVPR'22)}~\cite{liu2021video-swin}             & ImageNet-21K     & 84.0\\
    Uniformer-B~\textit{(ICLR'22)}~\cite{li2022uniformer}                & ImageNet-1K      & 84.5\\
    MoViNet-A6~\textit{(CVPR'21)}~\cite{kondratyuk2021movinets}          & ImageNet-21K     & 84.8\\
    MViTv1-B~\textit{(ICCV'21)}~\cite{fan2021multiscale}                 & None             & 83.8\\
    MViTv2-B~\textit{(CVPR'22)}~\cite{li2022improved}                    & None             & 85.5\\
    \midrule
    Video-FocalNet-B                                                     & ImageNet-1K      & \textbf{86.7}\\
    \bottomrule
\end{tabular}
}
\label{tab:k600}
\vspace{-.3cm}

\end{table}
\begin{table}[t!]

\centering
\small
\caption{Comparison with state-of-the-art methods on Something-Something-v2~\cite{goyal2017ssv2} dataset.}\vspace{-0.5em}
\setlength{\tabcolsep}{3mm}
\resizebox{0.95\columnwidth}{!}{
\begin{tabular}{lcc}
\rowcolor{Gray}
    \toprule
    Method                                                               & Pre-training     & Top-1\\
    \midrule
    SlowFast R50~\textit{(ICCV'19)}~\cite{feichtenhofer2019slowfast}     & ImageNet-21K     & 61.7\\
    TimeSformer-HR~\textit{(ICML'21)}~\cite{gberta2021timesformer}       & ImageNet-21K     & 62.5\\
    VidTR~\textit{(ICCV'21)}~\cite{zhang2021vidtr}                       & ImageNet-21K     & 63.0\\
    ViViT-L FE~\textit{(CVPR'21)}~\cite{arnab2021vivit}                  & ImageNet-21K     & 65.9\\
    MFormer-L~\textit{(NeurIPS'21)}~\cite{patrick2021keeping}            & ImageNet-21K     & 68.1\\
    MTV-B~\textit{(CVPR'22)}~\cite{shen2022mtv}                          & ImageNet-21K     & 67.6\\
    MTV-B (320p)~\textit{(CVPR'22)}~\cite{shen2022mtv}                   & ImageNet-21K     & 68.5\\
    Video-Swin-B~\textit{(CVPR'22)}~\cite{liu2021video-swin}             & Kinetics400      & 69.6\\
    Uniformer-B~\textit{(ICLR'22)}~\cite{li2022uniformer}                & Kinetics400      & 70.4\\
    MViTv1-B~\textit{(ICCV'21)}~\cite{fan2021multiscale}                 & ImageNet-21K     & 67.6\\
    MViTv2-B~\textit{(CVPR'22)}~\cite{li2022improved}                    & Kinetics400      & 70.5\\
    \midrule
    Video-FocalNet-B                                                     & Kinetics400      & \textbf{71.1}\\
    \bottomrule
\end{tabular}
}
\label{tab:ssv2}
% \vspace{-.3cm}

\end{table}
\begin{table}[t!]

\centering
\small
\caption{Comparison with state-of-the-art methods on Diving 48 V2~\cite{yingwei2018d48} dataset.}\vspace{-0.5em}
\setlength{\tabcolsep}{3mm}
\resizebox{\columnwidth}{!}{
\begin{tabular}{lcc}
\rowcolor{Gray}
    \toprule
    Method                                                                  & Pre-training                  & Top-1\\
    \midrule
    SlowFast R50~\textit{(ICCV'19)}~\cite{feichtenhofer2019slowfast}        & ImageNet-21K                  & 77.6\\
    TimeSformer-L~\textit{(ICML'21)}~\cite{gberta2021timesformer}           & ImageNet-21K                  & 81.0\\
    RSANet R50~\textit{(NeurIPS'21)}~\cite{kim2021relational}               & ImageNet-1K                   & 84.2\\
    VIMPAC~\textit{(arXiv'21)}~\cite{tan2022vimpac}                         & HowTo100M                     & 85.5\\
    BEVT~\textit{(CVPR'22)}~\cite{wang2021bevt}                             & Kinetics400                   & 86.7\\
    GC-TDN~\textit{(CVPR'22)}~\cite{yanbin2022gc}                           & ImageNet-1K                   & 87.6\\
    ORVIT Transformer~\textit{(CVPR'22)}~\cite{herzig2022objectregion}      & ImageNet-21K                  & 88.0\\
    TFCNET~\textit{(arXiv'22)}~\cite{zhang2022tfcnet}                       & ImageNet-1K                   & 88.3\\

    \midrule
    Video-FocalNet-B                                                        & Kinetics400                   & \textbf{90.8}\\
    \bottomrule
\end{tabular}}
\label{tab:d48}
% \vspace{-.3cm}

\end{table}
\begin{table}[!t]

\centering

\caption{Comparison on ActivityNet 1.3~\cite{heilbron2015anet} dataset.}\vspace{-0.5em}
\setlength{\tabcolsep}{4pt}
      \resizebox{\columnwidth}{!}{
\begin{tabular}{lccc}
\rowcolor{Gray}
    \toprule
    Method              & Pre-training     & Top-1 \\
    \midrule
    Video-Swin-B~\textit{(CVPR'22)}~\cite{liu2021video-swin}        & Kinetics400      & 88.5 \\
    \midrule
    Video-FocalNet-B    & Kinetics400      & \textbf{89.8} \\
    \bottomrule
\end{tabular}
}
\label{tab:anet}
\vspace{-0.3em}

\end{table}

\subsection{Experimental Setup and Protocols}
\label{sec:exp:experimental_setup}

\textbf{Datasets:} We report results for video action recognition on three large-scale datasets, Kinetics-400 (K400)~\cite{kay2017k400}, Kinetics-600 (K600)~\cite{carreira2018k600} and Something-Something-v2 (SS-v2)~\cite{goyal2017ssv2}. For each dataset, we train on the training set and evaluate on the validation set. K400 consists of $\sim$240k training and $\sim$20k testing videos across 400 classes. K600 consists of $\sim$370k training and 28.3k testing videos across 600 classes. SS-v2 consists of 169k training and 24.7k validation videos across 174 classes. For all three datasets, we report Top-1 accuracy and compare it against the state-of-the-art.

We additionally test our Video-FocalNet on the Diving-48 (D-48)~\cite{yingwei2018d48} and ActivityNet-1.3 (ANet-1.3)~\cite{heilbron2015anet} datasets. D-48 is a challenging dataset of diving actions consisting of $\sim$15000 training and $\sim$2000 testing samples. Actions are only differentiated by the subtle movement of the diver across the frames with the background being majorly constant. This means that the dataset requires robust temporal modeling for good performance. In fact, it has been shown by~\cite{wang2021bevt}, that disrupting (through random shuffling) or removing (by single frame evaluation) temporal information in this dataset can result in accuracy drops of up to $\sim 33.6\%$ and $\sim 70.2\%$ respectively. Alternatively, the ANet-1.3 dataset consists of untrimmed videos for action recognition tasks.

\textbf{Implementation Details:} For K400 and K600, we follow a similar training scheme to~\cite{li2022uniformer, li2022improved} and train for $120$ epochs with a linear warmup of $20$ epochs using the SGD optimizer. We linearly scale the learning rate by $LR \times \frac{batchsize}{512}$ where $LR = 0.1$ is the base learning rate. The spatial modules are initialized from the pretrained Imagenet-1K FocalNet~\cite{yang2022focal} weights, while the rest are randomly initialized. For augmentations, we follow a recipe similar to~\cite{fan2021multiscale} with some variations. To each clip, we apply a horizontal flip, Mixup~\cite{zhang2018mixup} ($\alpha$=0.8), and CutMix~\cite{yun2019cutmix}, each with a probability of 0.5. See detailed hyperparameters in \autoref{tab:training_hyperparameters}.

During training, we sample $T$ frames with a stride of $\tau$, denoted as $T \times \tau$~\cite{feichtenhofer2019slowfast}. For the spatial domain, we follow Inception~\cite{szegedy2015inception} and take a crop of $H \times W = 224 \times 224$, with input area selected within a scale of $[\min, \max] = [0.08, 1.00]$ and aspect ratio jitter between $3/4$ and $4/3$. During inference, we report results as an average across $N_{clip} \times N_{crops}$ where a total of $N_{clip}$ clips are uniformly sampled from the video, and for each video, $N_{crops}$ spatial crops are taken during inference. For K400 and K600 we use $4 \times 3$ for inference.
For SS-v2, D-48 and ANet-1.3 we follow the same training recipe as K400 and K600, with slight changes as followed by~\cite{liu2021video-swin, li2022uniformer, fan2021multiscale, li2022improved}. We initialize our model with the K400 pretrained weights. For augmentations, we don't use the random horizontal flip and infer on $1 \times 3$ views.

Additionally, owing to the large scale of the Kinetics-400~\cite{kay2017k400} and Kinetics-600~\cite{carreira2018k600} datasets, we preprocess the videos before starting to train. Following the guidelines of~\cite{2020mmaction2}, each video is first resized, with the shorter side resized to 256 pixels.

\subsection{Comparison with State-of-the-art}
\label{sec:exp:sota}

\textbf{Kinetics-400:} On the K400~\cite{kay2017k400} dataset, we report results for the Video-FocalNet-T, Video-FocalNet-S and Video-FocalNet-B variants, comparing against recent methods in \autoref{tab:k400}. Considering first the T and S variants, it can be seen that our method surpasses the equivalent Video-Swin Transformer~\cite{liu2021video-swin} variants by $1.0 \%$ and $0.8 \%$ respectively while reducing the TFLOPs by $25 \%$. Our larger base model, Video-FocalNet-B, surpasses the previous state-of-the-art Uniformer-B~\cite{li2022uniformer} and MViTv2-B~\cite{li2022improved} by $0.6 \%$ and $0.7 \%$ respectively, while maintaining comparable TFLOPs with MViTv2-B~\cite{li2022improved} and reducing TFLOPs by about $45 \%$ compared to Uniformer-B~\cite{li2022uniformer}.

\textbf{Kinetics-600:} On the K600~\cite{carreira2018k600} dataset, we report results for Video-FocalNet-B against recent methods in literature in \autoref{tab:k600}. Compared to the previous state-of-the-art MViTv2-B~\cite{li2022improved}, our Video-FocalNet-B achieves $1.2 \%$ higher performance. Our method using the ImageNet-1K initialization also surpasses previous methods pretrained on the larger ImageNet-21K dataset while maintaining much lower TFLOPs.

\textbf{Something-Something-v2:} On the SS-v2~\cite{goyal2017ssv2} benchmark we report results for Video-FocalNet-B and compare against state-of-the-art methods in \autoref{tab:ssv2}. On this temporally challenging benchmark, our method surpasses the previous state-of-the-art MViTv2-B~\cite{li2022improved} and Uniformer-B~\cite{li2022uniformer} by $0.6\%$ and $0.7 \%$ respectively. This strong performance shows that our method can effectively model the subtle temporal changes and dependencies in this challenging dataset.

\textbf{Diving-48:} On D-48~\cite{yingwei2018d48} we report our results and compare them against recent methods in literature in \autoref{tab:d48}. Video-FocalNet-B surpasses the previous state-of-the-art method TFCNET~\cite{zhang2022tfcnet} by $2.5\%$. This shows that our method can effectively model the temporal information even when using a small number of training samples.

\textbf{ActivityNet-1.3:} For ANet-1.3~\cite{heilbron2015anet}, results are presented in \autoref{tab:anet}. Our proposed Video-FocalNet outperforms the baseline Video-Swin (CVPR'22)~\cite{liu2021video-swin} model by a significant margin on the untrimmed video dataset. This demonstrates the efficacy of our method in localizing highlights and addressing the challenges posed by untrimmed videos. We appreciate your insightful suggestion and believe that evaluating our method on untrimmed video datasets further supports its potential in this challenging problem setting.

\begin{figure}[t]
    \centering
    \includegraphics[width=1\columnwidth]{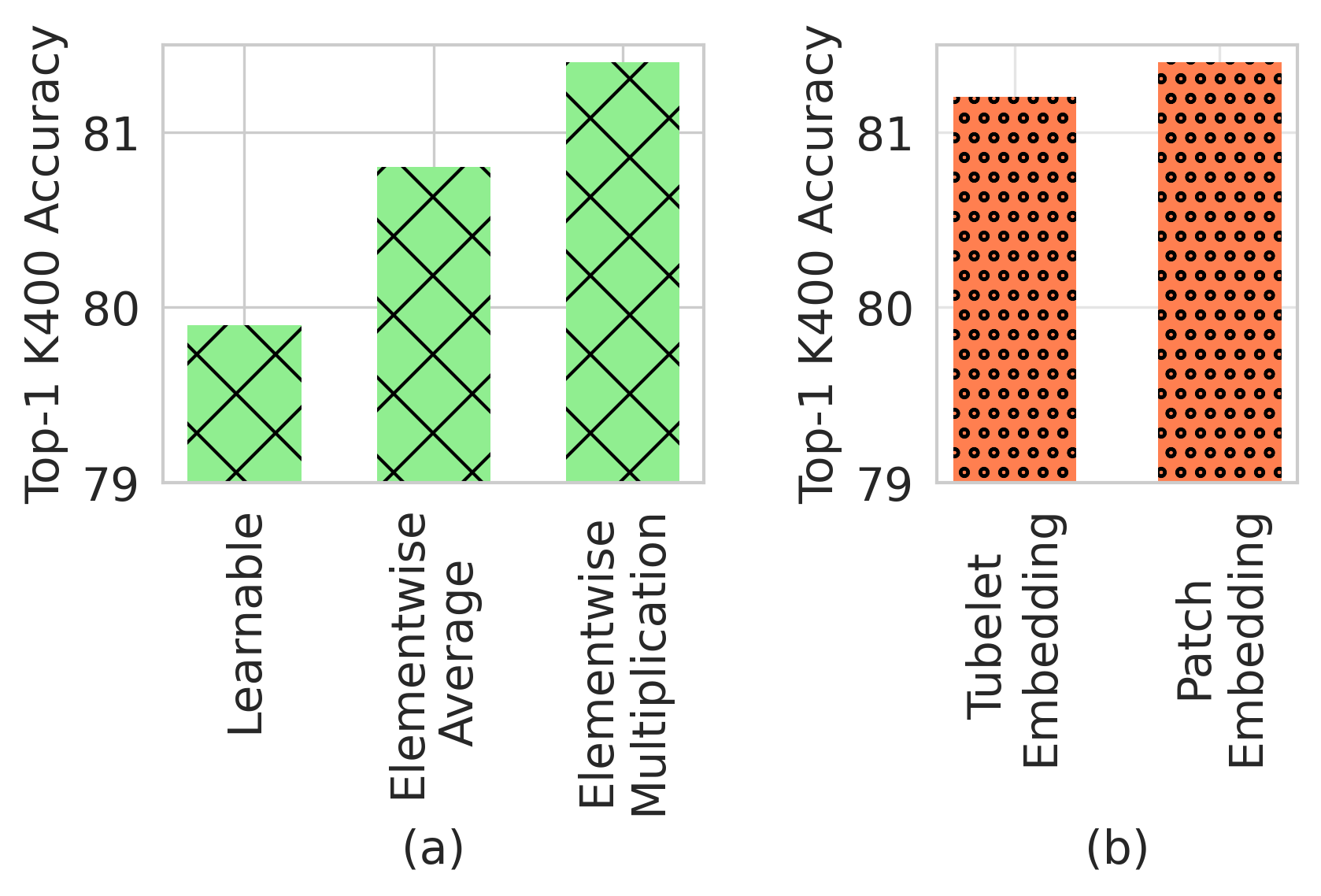}\vspace{-0.5em}
    \caption{\textbf{(a)} Ablation of various modulator-query spatio-temporal interaction methods for Video-FocalNet-S on Kinetics-400~\cite{kay2017k400} validation set. \textbf{(b)} Ablation for using patch vs tubelet embedding for Video-FocalNet-S on Kinetics-400~\cite{kay2017k400} validation set. Note that the number of frames is adjusted to ensure that the number of tokens is the same.}
    \label{fig:ablation} 
    \vspace{-0.3cm}
\end{figure}
\begin{figure*}[t]
\centering
        \begin{subfigure}[t]{0.3\textwidth}
        \centering
        \includegraphics[width=\linewidth]{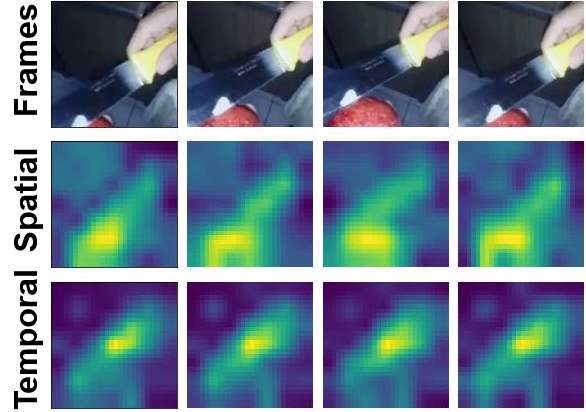}
        \caption{Class Label: "Cutting Apple"}
        \label{k400:1}
        \end{subfigure}\;\;
        \begin{subfigure}[t]{0.3\textwidth}
        \centering
        \includegraphics[width=\linewidth]{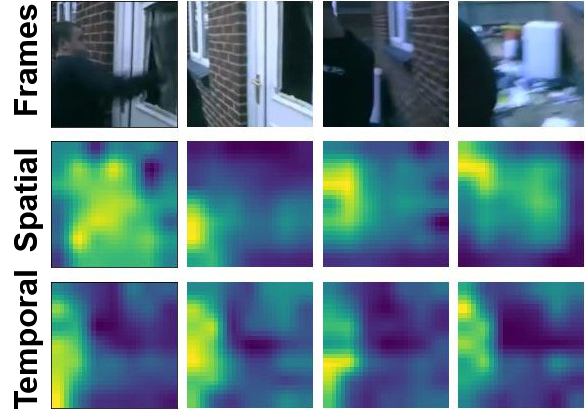}
        \caption{Class Label: "Smashing"}
        \label{k400:2}
        \end{subfigure}\;\;
        \begin{subfigure}[t]{0.3\textwidth}
        \centering
        \includegraphics[width=\linewidth]{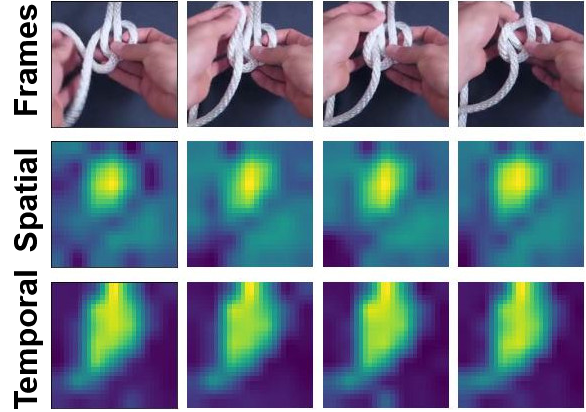}
        \caption{Class Label: "Tying Knot"}
        \label{k400:3}
        \end{subfigure}
    
        \begin{subfigure}[t]{0.3\textwidth}
        \centering
        \includegraphics[width=\linewidth]{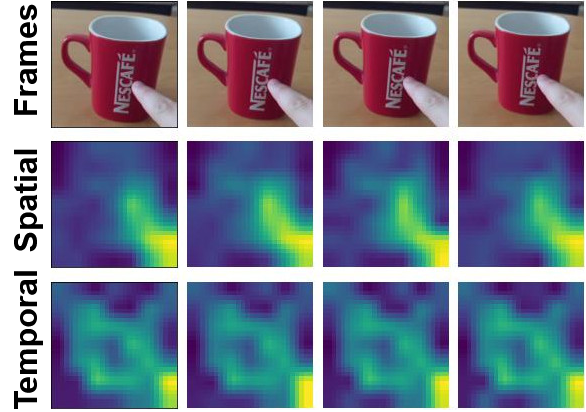}
        \caption{Class Label: "Pushing Something So That It Slightly Moves"}
        \label{ssv2:1}
        \end{subfigure}\;\;
        \begin{subfigure}[t]{0.3\textwidth}
        \centering
        \includegraphics[width=\linewidth]{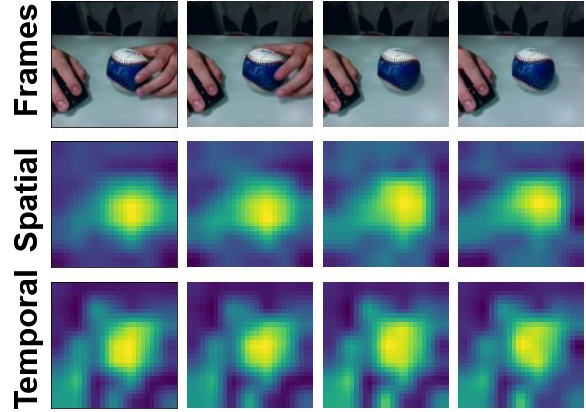}
        \caption{Class Label: "Putting Something On a Flat Surface Without Letting It Roll"}
        \label{ssv2:2}
        \end{subfigure}\;\;
        \begin{subfigure}[t]{0.3\textwidth}
        \centering
        \includegraphics[width=\linewidth]{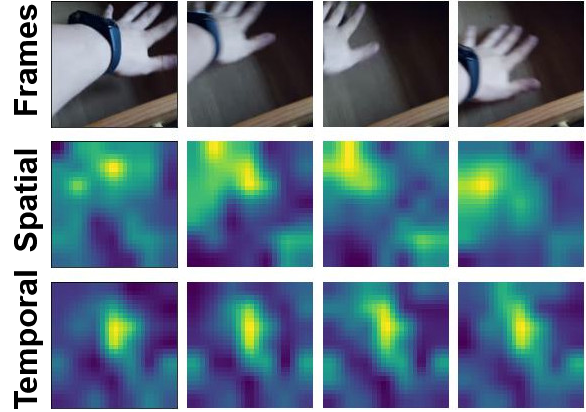}
        \caption{Class Label: "Showing That Something Is Empty"}
        \label{ssv2:3}
        \end{subfigure}

        \begin{subfigure}[t]{0.3\textwidth}
        \centering
        \includegraphics[width=\linewidth]{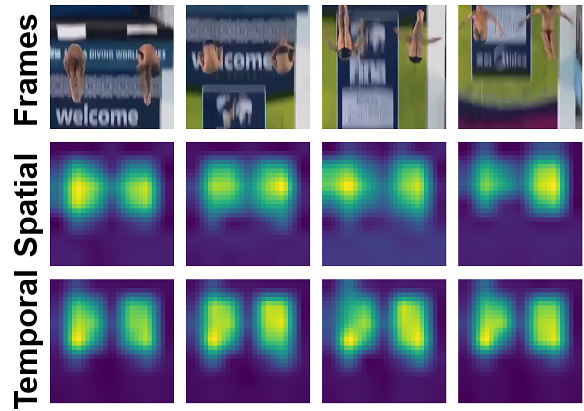}
        \caption{Class Label: ["Back", "25som", "15Twis", "PIKE"]}
        \label{d48:1}
        \end{subfigure}\;\;
        \begin{subfigure}[t]{0.3\textwidth}
        \centering
        \includegraphics[width=\linewidth]{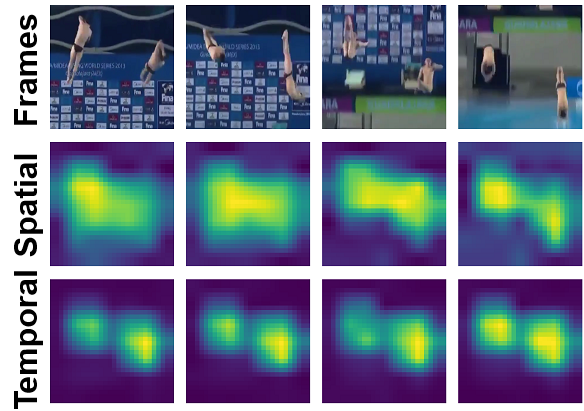}
        \caption{Class Label: ["Forward", "25som", "2Twis", "PIKE"]}
        \label{d48:2}
        \end{subfigure}\;\;
        \begin{subfigure}[t]{0.3\textwidth}
        \centering
        \includegraphics[width=\linewidth]{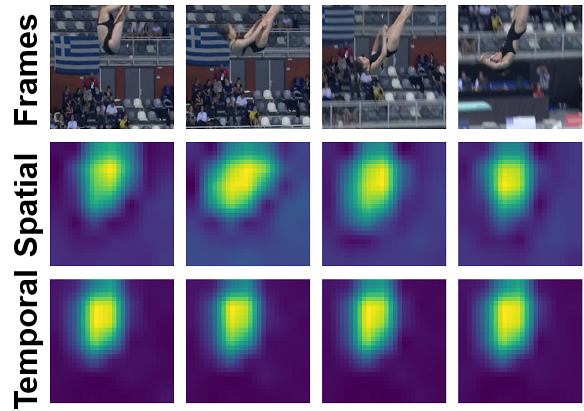}
        \caption{Class Label: ["Reverse", "Dive", "NoTwis", "PIKE"]}
        \label{d48:3}
        \end{subfigure}
        
        \caption{We visualize the spatial and temporal modulators for sample videos from Kinetics-600~\cite{carreira2018k600} (top row), Something-Something-V2~\cite{goyal2017ssv2} (middle row) and Diving-48~\cite{yingwei2018d48} (bottom row). Note how the temporal modulator fixates on the \emph{global} motion across frames while the spatial modulator captures \emph{local} variations. For example in \autoref{k400:1}, the temporal modulator specifically focuses on the point where the knife meets the apple, while the spatial modulator shifts focus from frame to frame based on the knife's position. For Diving-48 (bottom row),  we can see that the model can specifically fixate on the area where the action happens in each frame, regardless of the camera movement and small region of interest. More interestingly, the temporal modulator can separate the two regions of action in \autoref{d48:1} and \autoref{d48:2}.}

    \label{fig:vis}
\vspace{-.3cm}
\end{figure*}

\subsection{Ablations}
\label{sec:exp:ablations}

In this section, we present an ablative analysis of various choices in our final design. Note that all ablations are performed using the Video-FocalNet-S variant on K400 using the same training settings as mentioned in \autoref{sec:exp:experimental_setup}.

\textbf{Modulator Fusion Method:} Since we propose a two-stream spatio-temporal focal modulation design, we end up with two modulators, one each for the spatial and temporal branches respectively, that need to be fused with the query tokens. We evaluate various fusion methods to see which works best. \autoref{fig:ablation} \textbf{(a)} shows the comparison of three fusion techniques which include simple averaging, elementwise multiplication, and a learnable projection layer. We find that elementwise multiplication gives the best performance.

\textbf{Patch Embedding vs Tubelet Embedding:} Many recent works~\cite{liu2021video-swin, arnab2021vivit, shen2022mtv} propose encoding a tubelet of $T \times H \times W \times 3$, with $T=2$, into a single token rather than patch embedding with $T=1$. We evaluate this design choice for our model and find that a simple patch embedding works better for us, as shown in \autoref{fig:ablation} \textbf{(b)}.

\textbf{Visualizations:} We visualize the spatial and temporal modulators for sample videos across two datasets, K600 and SS-V2 in \autoref{fig:vis}. We note that our modulators focus on the salient parts and essential dynamics of the video which are relevant to the end task. The spatial modulator tends to shift to the \emph{local} spatial changes in individual frames, while the temporal modulator fixates to the \emph{global} region across frames where the majority of the motion happens. 
%-------------------------------------------------------------------------

\section{Conclusion}
To learn spatio-temporal representations that can effectively model both local and global contexts, this paper introduces Video-FocalNets for video action recognition tasks. This architecture is derived from focal modulation for images and can effectively model both short- and long-term dependencies to learn strong spatio-temporal representations. We extensively evaluate several design choices to develop our proposed Video-FocalNet block. Specifically, our Video-FocalNet uses a parallel design to model hierarchical contextualization by combining spatial and temporal convolution and multiplication operations in a computationally efficient manner. Video-FocalNets are more efficient than transformer-based architectures which require expensive self-attention operations. We demonstrate the effectiveness of Video-FocalNets via evaluations on five representative large-scale video datasets, where our approach outperforms previous transformer- and CNN-based methods.

%-------------------------------------------------------------------------

{\small
\bibliographystyle{ieee_fullname}
\bibliography{egbib}
}

\end{document}